\documentclass[letterpaper, 10 pt, conference]{ieeeconf}  

\IEEEoverridecommandlockouts 
\usepackage[T1]{fontenc}

\usepackage{amsmath}
\usepackage{amssymb}
\usepackage{tabularx}
\usepackage{booktabs}
\usepackage{soul, color}
\usepackage{url}
\usepackage{graphicx}
\usepackage{caption}
\usepackage{anyfontsize}
\usepackage{subcaption}
\usepackage{cite}
\usepackage{xspace}
\usepackage{comment} 

\newcommand{\ie}{\textit{i.e.}, }
\newcommand{\eg}{\textit{e.g.}, }
\newcommand{\etal}{\textit{et al.}\xspace}

\DeclareCaptionFont{xxviii}{\fontsize{9}{9}\selectfont}
\makeatletter
\DeclareRobustCommand{\change}{%
  \@bsphack
  \leavevmode 
  \color{red} 
  \@esphack
}
\DeclareRobustCommand{\stopchange}{%
  \@bsphack
  \normalcolor
  \@esphack
}
\makeatother

\title{\LARGE \bf
Towards Effective Visual-Inertial SLAM with Passive-Only Sensors for Low-Cost Autonomous Underwater Vehicles}

 \author{Grant Schwidder, David Widhalm, and Junaed Sattar$^{*}$
 \thanks{$^{*}$This work was supported in part by the National Science Foundation Grant \#2220956, and the Science, Mathematics, Research for Transformation (SMART) Program.
 The authors are with the Department of Computer Science \& Engineering, University of Minnesota--Twin Cities, Minneapolis, MN 55455, USA
 {\tt\small \{schwi488, widha008, junaed\}@umn.edu}}%
 }

\begin{document}

\maketitle
\thispagestyle{empty}
\pagestyle{empty}

\begin{abstract}
Improvements to Visual-Inertial Simultaneous Localization and Mapping (VI-SLAM) for low-cost autonomous underwater vehicles (AUVs) are critical for transitioning advanced marine robotics from specialized labs to broader research and hobbyist applications. 
While high-end AUVs typically rely on expensive sensor suites --- such as Doppler Velocity Logs (DVLs) and Ultra-Short Baseline (USBL) systems --- this work demonstrates that robust, high-quality navigation is achievable using a sub-\$$10,000$(USD) platform equipped only with inexpensive consumer-grade sensors. By leveraging a similarly priced, open-source AUV, we evaluate the performance of stereo cameras, Micro-electromechanical System (MEMS)-based IMUs, and depth sensors in a fully unconstrained 6-degree-of-freedom (6-DOF) underwater environment. We analyze the efficacy of off-the-shelf SLAM packages and propose optimizations for sensor fusion to mitigate the visual and physical challenges of untethered underwater operation. Our results prove that a usable SLAM solution can be accessible to the masses, providing a benchmark for expectations in demanding, real-time maritime missions without the financial barrier of industrial-grade hardware.

\end{abstract}

\section{Introduction}
\label{sec:introduction}
Hobbyist and low-cost underwater robotics often rely on off-the-shelf hardware and proprietary Application Programming Interfaces (APIs), which can impose significant constraints on system development. 
For researchers with limited resources, achieving autonomous control and motion is particularly difficult due to the absence of Global Positioning System (GPS) signals in maritime (and in particular, underwater) environments. 
While industrial-grade Autonomous Underwater Vehicles (AUVs) overcome this using expensive active acoustic sensors -- such as Doppler Velocity Logs (DVLs)~\cite{kimRealTimeVisualSLAM2013} or Ultra-Short Baseline (USBL)~\cite{wadhamsOperationalHistoryDevelopment2019} systems -- the cost of these components often exceeds the entire budget of an accessible platform.

\begin{figure}[t]
\vspace{3mm}
    \begin{subfigure}{\columnwidth}
        \includegraphics[width=\columnwidth]{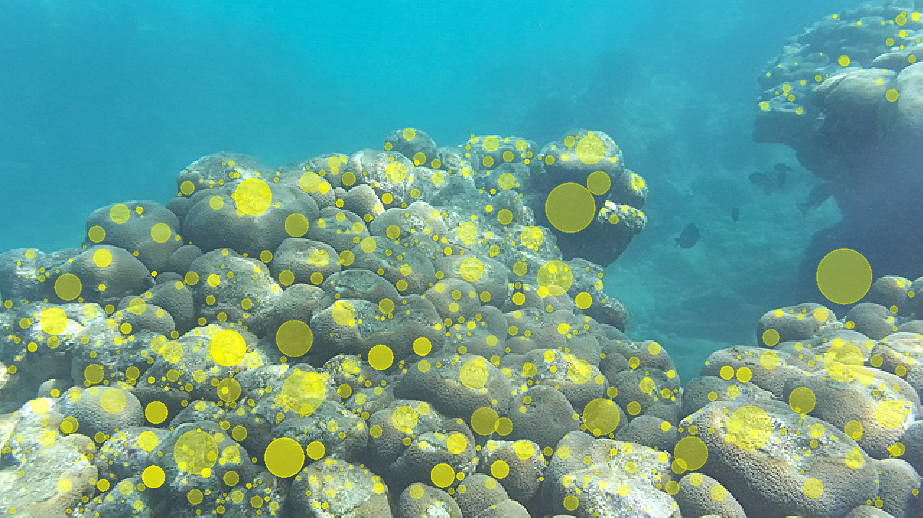} 
        \caption{Camera image from AUV showing features in ocean reef}
        \label{fig:reefCamera} 
    \end{subfigure}
    \begin{subfigure}{\columnwidth}
      \centering
         \includegraphics[width=\columnwidth]{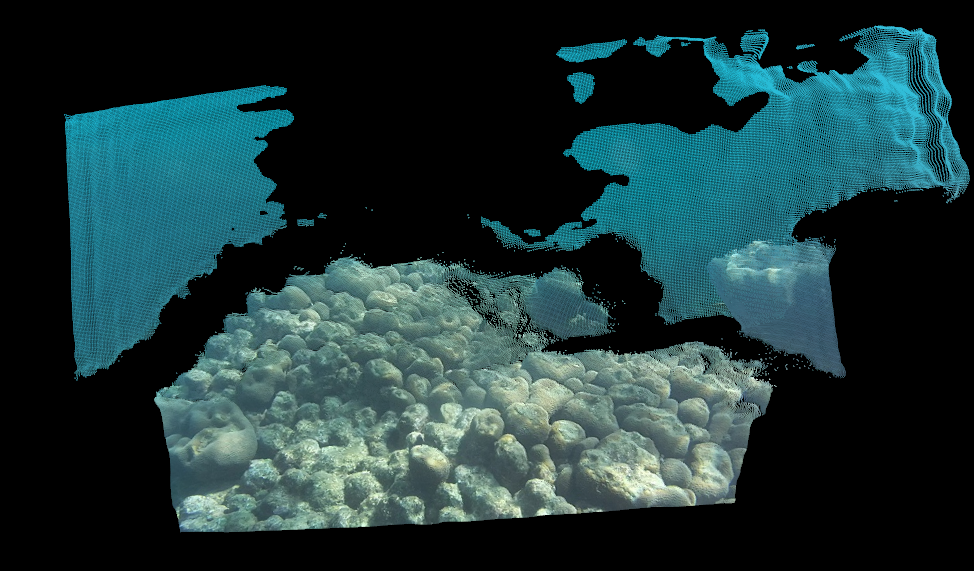}
        \caption{Sample point cloud of coral reef constructed from AUV data}
        \label{fig:reefPointCloud}
    \end{subfigure}
\caption{A snapshot of VI-SLAM running on board our low-cost AUV on an ocean reef. Fig.~\ref{fig:reefCamera} shows actual images acquired by the on-board stereo camera and visual features used for SLAM. Fig.~\ref{fig:reefPointCloud} shows a point cloud representation of the reef reconstructed aftwerward.}
\label{fig:reef}
\end{figure}

To ensure a robot remains accessible to field researchers and other end users for tasks like environmental monitoring and local mapping, the hardware must be easily deployable and financially viable~\cite{johnson-robersonHighResolutionUnderwaterRobotic2017}. 
This study exploits a platform designed with purely passive, low-cost sensors to reduce weight, power consumption, and environmental impact. 
By avoiding tethers, base stations, and high-cost external equipment, the system remains a standard tool for ``masses-accessible'' underwater robotics. 
However, relying on a Visual SLAM (VSLAM) (\eg~\cite{leonardiUVSUnderwaterVisual2023}) approach presents significant challenges, as underwater conditions are often visually degraded and physically demanding.

The navigation suite for this work consists of a stereo camera with an integrated Inertial Measurement Unit (IMU), a secondary Inertial Navigation System (INS), and a pressure-based depth sensor. 
Because these sensors publish at varying rates and exhibit different error profiles -- specifically significant IMU drift -- a robust fusion strategy is required. 
Using the Robot Operating System 2 (ROS 2) framework~\cite{ROSGettingStarted}, Extended Kalman Filter (EKF) nodes were implemented to synchronize and fuse these data streams. 
While initial out-of-the-box SLAM performance yielded large mapping errors, we demonstrate that algorithmic optimizations and multi-sensor fusion can compensate for the lack of high-end active sensors.

By implementing dense representations and refined motion compensation, this work provides a significantly improved SLAM system capable of maintaining accuracy and achieving loop closure even during rapid, unconstrained 6-degree-of-freedom (6-DOF) rotational motions. 
These improvements were achieved through software and algorithmic tuning alone, requiring no additional hardware expenditures.

This work demonstrates that:
\begin{itemize}
    \item Fully Passive SLAM is viable for unconstrained 6-DOF underwater motion on budget-constrained platforms, and
    \item Precise algorithmic selection and sensor fusion are more critical than hardware cost when operating in degraded underwater environments.
\end{itemize}

\section{Related Work}
\label{sec:related_works} 
The landscape of underwater navigation is traditionally bifurcated between high-cost reliability and low-cost, but often experimental, fragility. 
This section explores the current state of maritime Simultaneous Localization and Mapping (SLAM) and the barriers to 6-DOF autonomy.

Standard underwater navigation relies heavily on the fusion of Doppler Velocity Logs (DVLs), pressure sensors, and high-grade Inertial Navigation Systems (INS).
Recent state-of-the-art frameworks, such as those proposed in~\cite{panVISORobustUnderwater2026}, demonstrate high-precision 6-DOF localization through tightly coupled acoustic-visual-inertial systems. 
While robust, these systems typically utilize DVLs to constrain the inherent drift of accelerometers. 
However, the hardware cost for a maritime-grade DVL often exceeds \$$15,000$, creating a significant financial barrier for research-grade or hobbyist Autonomous Underwater Vehicles (AUVs)~\cite{jalaliMaritimeTechnologyAttention2025}.
 
VSLAM is a well-studied topic and finds significant use in ground, air, and water vehicles. 
Barros \etal~\cite{macariobarrosComprehensiveSurveyVisual2022} produce a survey of VSLAM algorithms which is focused on above-water deployments. 
Monocular VSLAM systems are popular as the systems are small and light, Leonardi \etal describe a recent camera only system~\cite{leonardiUVSUnderwaterVisual2023}. 
Kim \etal describe an acoustic monocular SLAM in 2013~\cite{kimRealTimeVisualSLAM2013}, more recently Xu \etal and Rahman \etal propose multi-sensor VSLAM algorithms with acoustic sensors~\cite{xuAQUASLAMTightlyCoupled2025,rahmanSVIn2MultisensorFusionbased2022}. 

VSLAM offers a cost-effective alternative to acoustic positioning, yet it faces unique environmental degradations including light scattering, ``marine snow,'' and low-texture seafloors. 
While algorithms like ORB-SLAM3~\cite{camposORBSLAM3AccurateOpenSource2021} have shown promise in clear-water conditions, they frequently suffer from ``IMU under-excitation'' during the slow, steady-state motions typical of AUV survey missions. 
Most low-cost visual approaches remain tethered or require human-in-the-loop correction to compensate for rapid 6-DOF perturbations, particularly in turbid waters~\cite{ferreraRealTimeMonocularVisual2019}.

Monocular systems often rely on feature extraction~\cite{mur-artalORBSLAMVersatileAccurate2015,mur-artalORBSLAM2OpenSourceSLAM2017,camposORBSLAM3AccurateOpenSource2021} or neural networks~\cite{zhuNICERSLAMNeuralImplicit2024,yangMSGSSLAMMonocularSemantic2025} to determine scene depth. 
Underwater scenes are frequently feature poor, and the computational resources of the AUV need to be split between many tasks, not just mapping. 
Thus the overhead of additional networks for odometry and mapping is excessive for this use-case. 
Early stereo VSLAM underwater without additional sensors show significant drift even when traveling at a consistent depth~\cite{bonin-fontStereoSLAMRobust2015} while the aforementioned works of Xu \etal and Rahman \etal rely on active acoustics for additional accuracy. 
Johnson \etal~\cite{johnson-robersonHighResolutionUnderwaterRobotic2017} describe a scenario of mapping an underwater archaeological site with the associated challenges of underwater VSLAM in shallow conditions with an AUV using active acoustics.

There remains a critical research gap for untethered, unconstrained 6-DOF platforms operating within a sub-\$$10,000$ budget. 
Current literature often presents a binary choice: expensive, robust acoustic suites or low-cost, fragile visual systems. 
This work occupies the middle ground, demonstrating that through specific algorithmic tuning in the Robot Operating System 2 (ROS 2) framework and the fusion of consumer-grade stereo-inertial and depth data, an acceptable SLAM solution is achievable for field research and environmental monitoring without the need for industrial-grade active sensors.

\section{Methodology}
\label{sec:methodologies}
Here, we describe the proposed approach on our low-cost AUV, starting with the hardware description, and subsequently describing the motion compenstion and sensor fusion strategies we adopted. 

\subsection{Hardware and Systems Overview} 
\label{sec:hardware}
The navigation suite uses a stereo ZED Mini camera featuring an integrated IMU~\cite{ZEDMiniStereo}, a MicroStrain 3DM-CV7 IMU~\cite{3DMCV7INS}, and a BlueRobotics MS5837 depth sensor~\cite{BarHighResolutionDepth}. 
The ZED camera contains a baseline of $63$mm, allowing for an estimate of up to $15$m of depth \cite{ZEDMiniStereo}, and offers a Software Development Kit (SDK) that contains many different calculations for state estimation along with associated ROS2 interfaces. 
These allow for the use of Visual Inertial Odometry (VIO), using both stereo camera data and IMU data to create an odometry estimate. 
The ZED SDK has two main odometry estimation methods that can be used out of the box:

\begin{itemize}
    \item \textbf{GEN 1} Estimates motion by using a dense representation of the environment derived directly from the depth map. This method is highly robust in featureless environments as it does not rely on the extraction of distinct geometric features.
    
    \item \textbf{GEN 3} Estimates motion by extracting and tracking high-quality visual features across frames. Unlike GEN 1, it maintains a map of its environment as it traverses, and preforms loop closures and global corrections. While it offers the highest accuracy in structured or well-lit environments, its reliance on learned spatial features can be less predictable in highly turbid or visually repetitive underwater conditions. 
\end{itemize}
Given the consistent lack of quality features in real-world underwater settings, GEN 1 was used as the method for visual odometry estimation. 
The evaluations constituted of unconstrained motion in repetitive environments, where GEN 1 outperformed GEN 3 in all of our experiments which is expounded upon in Sec.~\ref{sec:experiments}.

\subsection{Motion Compensation and Coordinate Frames} 
\label{sec:motion_comp}
\begin{figure}[!ht]
    \centering
    \begin{subfigure}[b]{0.32\textwidth}
        \centering
        \includegraphics[width=\textwidth]{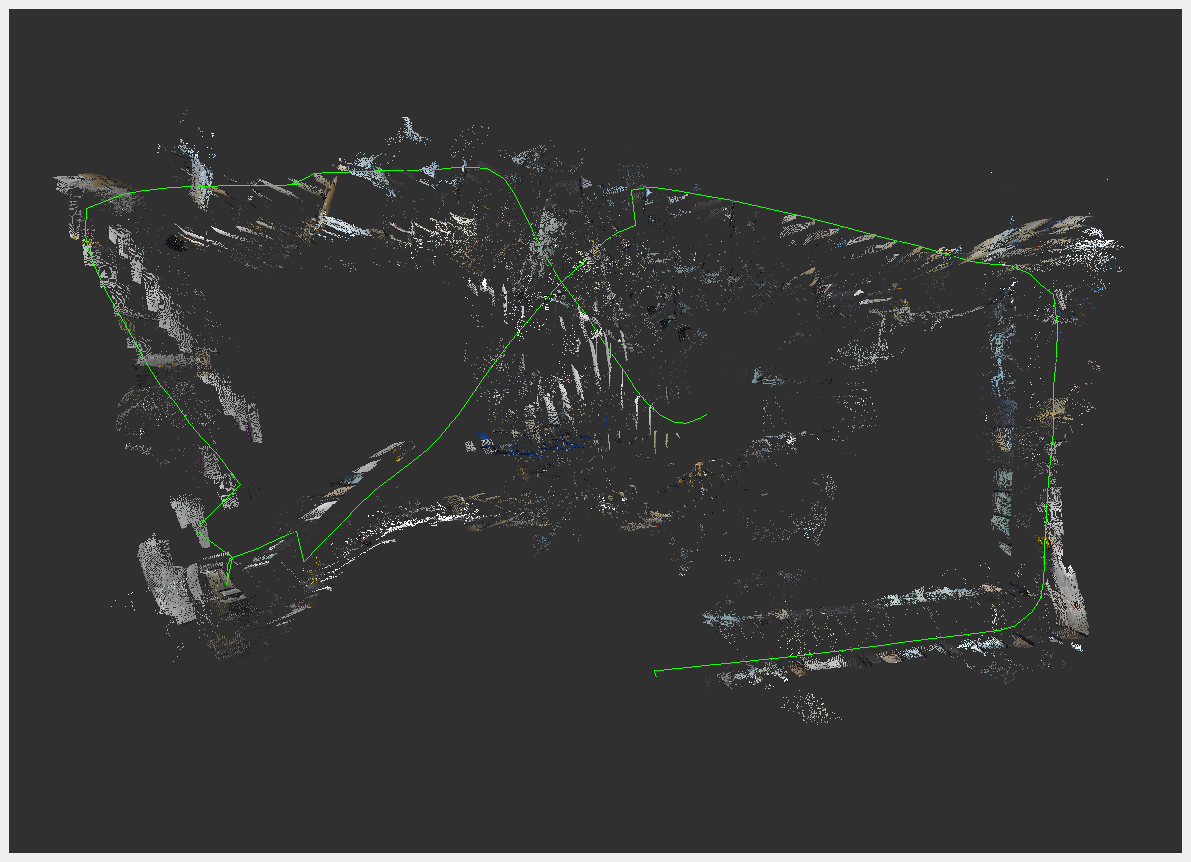}
        \caption{Absolute Pose Mode}
        \label{fig:abs_odom}
    \end{subfigure}
    \hfill
    \begin{subfigure}[b]{0.32\textwidth}
        \centering
        \includegraphics[width=\textwidth]{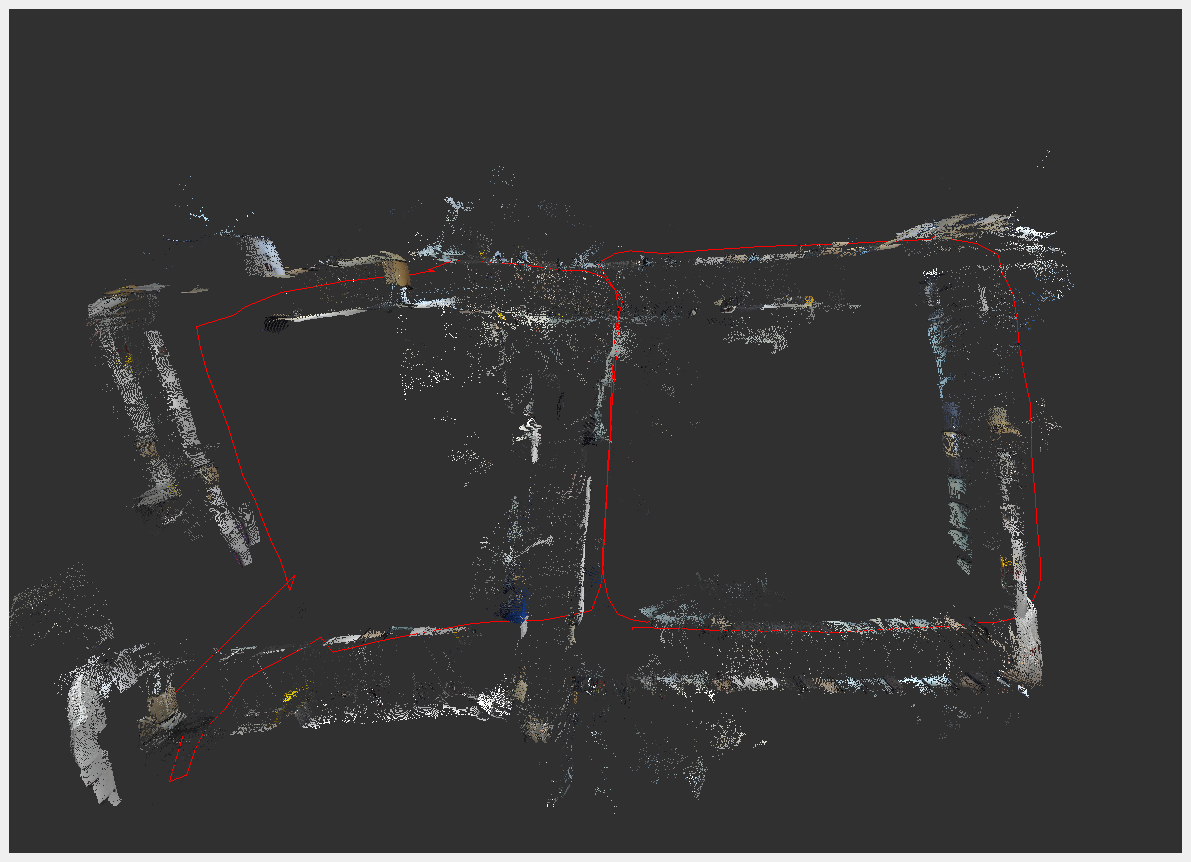}
        \caption{Differential Mode}
        \label{fig:diff_odom}
    \end{subfigure}
    \hfill
    \begin{subfigure}[b]{0.32\textwidth}
        \centering
        \includegraphics[width=\textwidth]{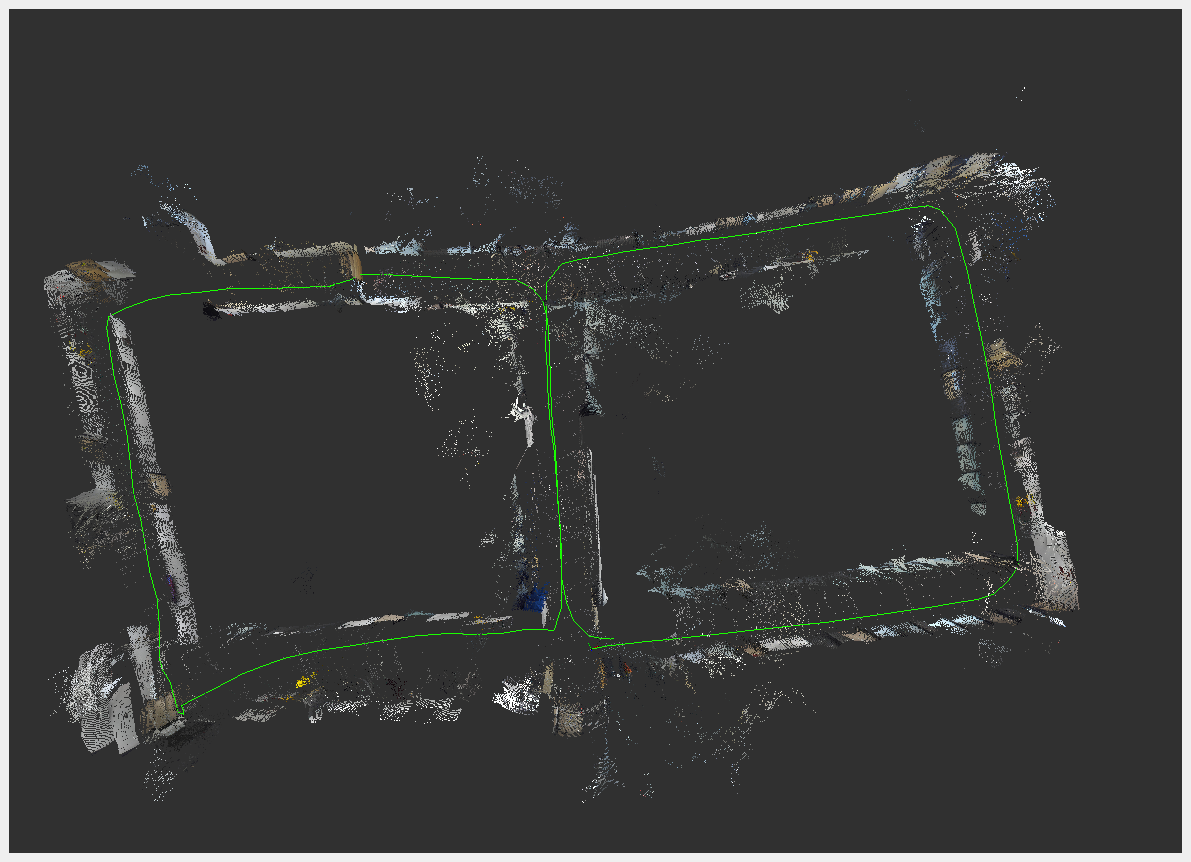}
        \caption{Full EKF Fusion}
        \label{fig:ekf_odom}
    \end{subfigure}
    \caption{Evolution of tracking performance: (a) Absolute VIO showing significant shearing; (b) Differential integration improving consistency; and (c) Full EKF fusion incorporating IMU and pressure data to eliminate drift and handle visual degradation. All implementations are running with RTAB-Map loop closures enabled.}
    \label{fig:odom_evolution}
\end{figure}

\begin{figure}[t]
    \vspace{2.5mm}
    \centering
    \includegraphics[width=\columnwidth]{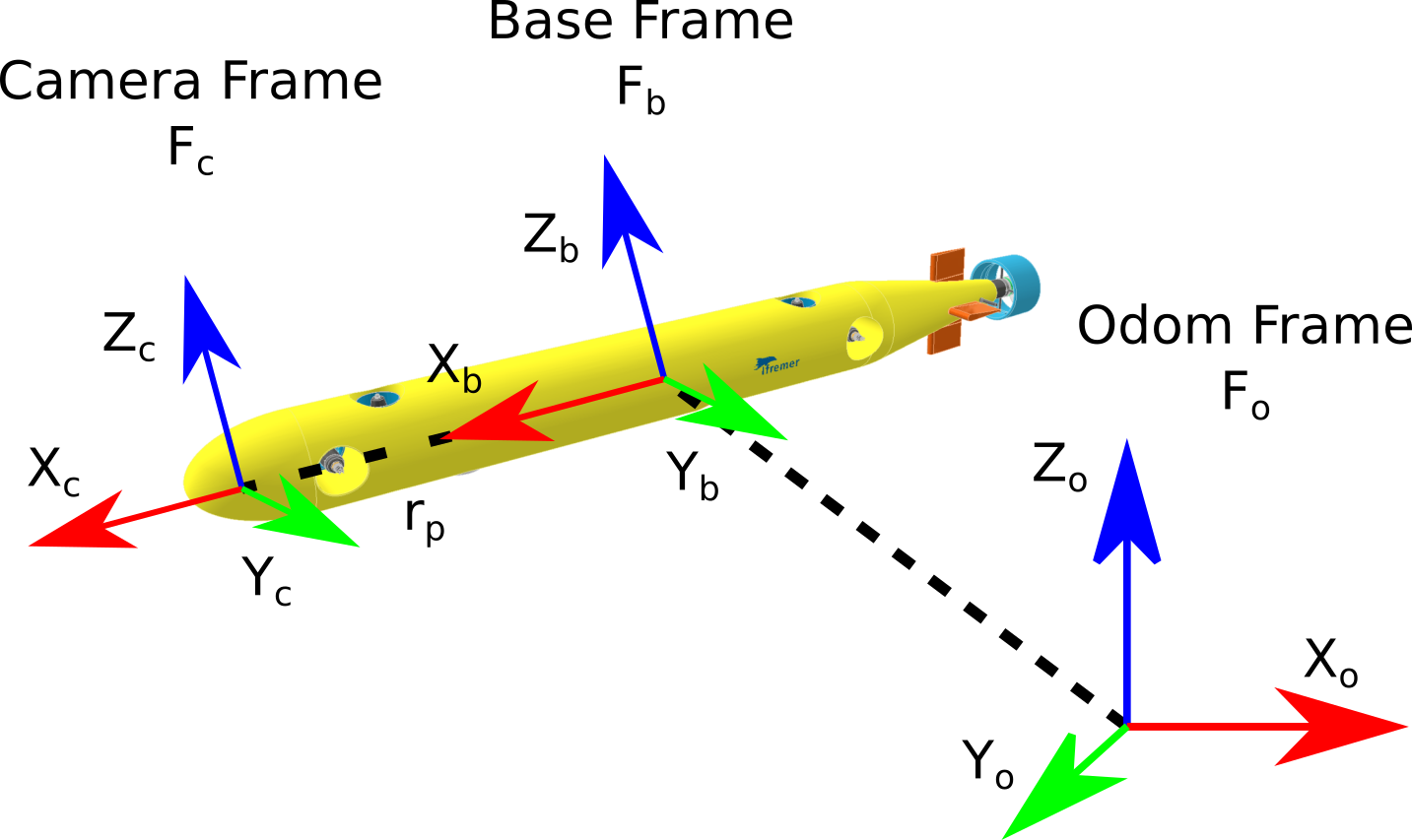}
    \caption{Layout of the coordinate frame configuration of the AUV, omitting the world frame. The $r_p$ vector shows the distance used in the Transport Theorem.}
    \label{fig:coordinate_frames}
\end{figure}

The ZED visual inertial odometry produces its results in absolute position (\ie Odom Frame, Fig.~\ref{fig:coordinate_frames}) values along with its current orientation. 
This absolute estimate quickly diverges (Fig.~\ref{fig:abs_odom}) during turns with a large angular velocity, since the measured orientation becomes inaccurate. 
With this poor orientation measurement, the absolute position would contain unbounded absolute error. 
To mitigate this, relative displacements were derived from the position estimates which controlled the error.

To address orientation inaccuracies during high-velocity yaw maneuvers (which lead to unbounded absolute position errors), we derived relative displacements rather than relying on raw absolute estimates. 
As linear velocity measured in the camera frame includes components from angular velocity, the Transport Theorem~\cite{alma9954828420001701} was applied to transform measurements from the camera frame into the robot's base frame (Fig.~\ref{fig:coordinate_frames}):

\begin{equation}
\mathbf{v}_{base} = \mathbf{R}_{cam}^{base} \mathbf{v}_{cam} - (\boldsymbol{\omega}_{base} \times \mathbf{r}_{p}),
\label{eq:transport_theorem}
\end{equation}


\subsection{Sensor Fusion Strategy} 
\label{sec:fusion_strategy}
Solely relying on ZED visual data for odometry encounters many failure cases. 
To address the drift observed in pure VIO, sensor data was fused using an Extended Kalman Filter (EKF)~\cite{smithAPPLICATIONSTATISTICALFILTER}, implemented using the ROS2 \verb|robot_localization| framework~\cite{mooreGeneralizedExtendedKalman2015}.
This allowed for the integration of the external pressure-based depth sensor and a secondary IMU to maintain state estimation during periods of visual degradation.
The framework provides a robust EKF implementation using ROS2, and also allows for asynchronous sensor data to be received and used by the filter, exposing many parameters to fine-tune during the state estimation process.
\vspace*{+5mm}
\begin{table*}[!t]
\centering
\vspace*{+5mm}
\caption{EKF configuration: active state variables per sensor}
\resizebox{\textwidth}{!}{
\begin{tabular}{l | ccc | ccc | ccc | ccc | ccc}
\hline
\textbf{Sensor} & \multicolumn{3}{c|}{\textbf{Pose}} & \multicolumn{3}{c|}{\textbf{Orientation}} & \multicolumn{3}{c|}{\textbf{Linear Vel. ($v$)}} & \multicolumn{3}{c|}{\textbf{Angular Vel. ($\omega$)}} & \multicolumn{3}{c}{\textbf{Acceleration ($a$)}} \\
 & $x$ & $y$ & $z$ & $\phi$ & $\theta$ & $\psi$ & $\dot{x}$ & $\dot{y}$ & $\dot{z}$ & $\dot{\phi}$ & $\dot{\theta}$ & $\dot{\psi}$ & $\ddot{x}$ & $\ddot{y}$ & $\ddot{z}$ \\ \hline
ZED VIO         & \checkmark & \checkmark & \checkmark & & & & & & & & & & & & \\
ZED IMU         & & & & \checkmark & \checkmark & \checkmark & & & & \checkmark & \checkmark & \checkmark & & &\\
2nd IMU        & & & & \checkmark & \checkmark & \checkmark & & & & \checkmark & \checkmark & \checkmark & \checkmark & \checkmark & \checkmark \\
Pressure Sensor & & & \checkmark & & & & & & & & & & & & \\ \hline
\end{tabular}
}
\label{tab:sensor_config}
\end{table*}

The acceleration from the MicroStrain IMU on the AUV allows for dead reckoning to be used when the visual odometry loses track. 
Covariance of the visual odometry is used as a measure of fit for the estimation. 
When the covariance for the visual odometry becomes larger than a predefined threshold, dead reckoning is used instead as it is expected to be more accurate. 
The orientation and the angular velocity from this IMU is fused into the EKF to improve the estimated state. 
The ZED IMU is used as a secondary reference, as it is less accurate and has a lower update rate. 
By utilizing the relative changes from the ZED sensor rather than its absolute orientation, the filter fuses this data with the absolute measurements from the primary MicroStrain IMU without inducing state oscillation. 
This prevents conflicting state updates between the two IMU sensors.

IMU performance was improved by using a static startup calibration period on the AUV. 
During the initial startup, the IMU Bias and local gravity vectors are measured and saved for removal from subsequent measurements. 
This improves the IMU performance by both accounting for the bias which can change on each power cycle, but also matches the local gravity vector instead of relying on a predefined value in the IMU. 
While both effects are small (less than $1$\% of of the gravity vector), they do improve the measurements of the IMU. 
The cost is ensuring that the AUV does not move during the first few seconds after it powers on.

The pressure sensor is used to give an absolute estimate of the $Z$ position, as measured from sea-level using standard pressure (STP). 
This significantly reduces drift accumulated from error in the visual odometry and IMU acceleration. 
This configuration successfully mitigated error in areas where the camera alone would fail (Fig.~\ref{fig:ekf_odom}). 
The specific state variables contributed by each sensor to the EKF are detailed in Table~\ref{tab:sensor_config}.

This fused state estimate is then integrated with RTAB-Map, an Open-Source Visual SLAM Library for Large-Scale and Long-Term Online Operation~\cite{labbeRTABMapOpenSourceLidar2019}. 
RTAB-Map utilizes the stereo images and depth data from the ZED camera in conjunction with the EKF’s odometry to perform loop closure detection and graph-based optimization, effectively neutralizing long-term odometry drift~\cite{labbeAppearanceBasedLoopClosure2013}. 
The system also implements memory management, determining what nodes stay in working memory, and moving inactive nodes to the disk, ensuring long term mapping is possible with memory constraints, allowing real time deployment on the Jetson Orin aboard the AUV.

This approach isolates the high-frequency odometry from the globally optimized SLAM graph by utilizing the standard ROS coordinate frame hierarchy ($map \to odom \to base\_link$)~\cite{REP105Coordinate}. 

The state estimation is published under the \verb|odom| frame, which is local and continuous, containing no global correction. 
This is less accurate, but can be used for real time correction in closed loop controllers. 
The optimized graph is published under the \verb|map| frame, which acts as the world frame, enabling its use for precise path planning. 
This also allows the SLAM backend to be flexible and modular in nature, as different packages could be implemented instead of RTAB-Map. 
\section{Experiments}
\label{sec:experiments}

The implementation of SLAM was evaluated in various operational environments with increasing complexity. 
These environments were specifically chosen to challenge the robustness of passive sensors onboard against dynamic lighting, perceptual aliasing, and feature scarcity. 
Experiments began with a rectilinear indoor office setting, then a shallow pool, and finally an open-ocean environment. 
The first setting, the office, consisted of interconnected hallways where the AUV followed a figure-eight trajectory to facilitate multiple loop-closure opportunities.

\subsection{Hallway Evaluation}
\label{sec:eval_hallway}
These hallways featured repetitive layouts and numerous windows, which present a challenge to the loop-closure back end. Such visual constraints test whether the system can maintain reliable odometry and successful loop closures despite perceptual aliasing and lighting interference. To assess the robustness of the Kalman Filter-based state estimation, two distinct motion profiles were compared: constant-radius turns and high angular-velocity turns. This allowed for a direct analysis of how sudden rotational changes impact mapping drift and filter convergence. Furthermore, it provided a framework to assess the IMU's capacity to mitigate visual odometry failures by maintaining accurate orientation estimates when camera-based tracking became unreliable.

\begin{figure}[ht]
 \vspace*{+3mm}
    \begin{subfigure}{\columnwidth}
        \includegraphics[width=\columnwidth]{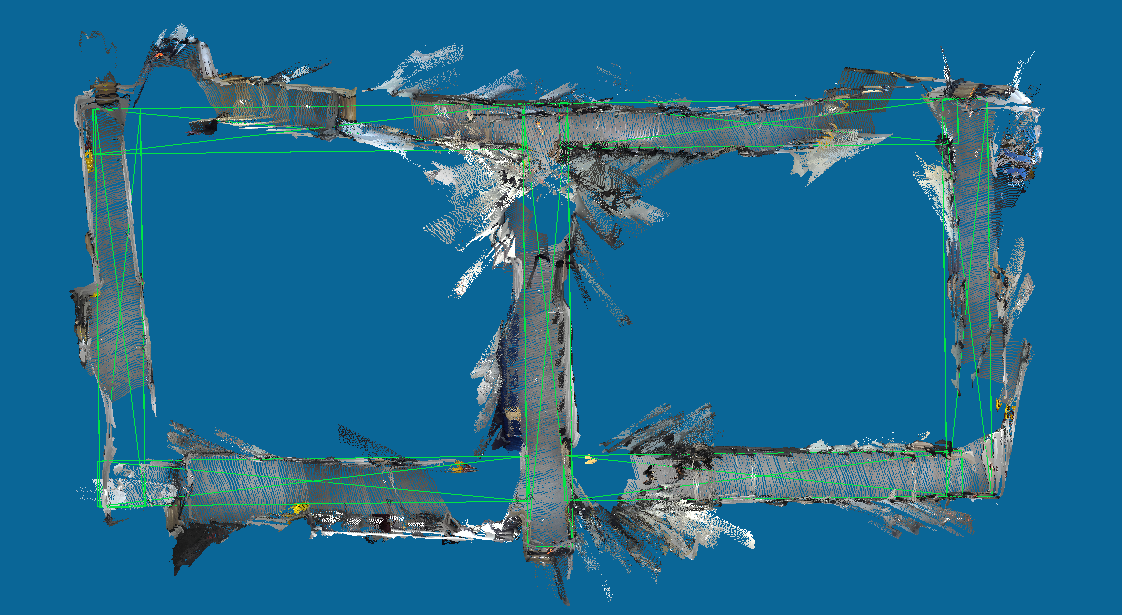}
        \caption{Map with constant radius turns overlaid with ground truth wire frame.}
        \label{fig:placeholder}  
    \end{subfigure}
    \begin{subfigure}{\columnwidth}
      \centering
        \includegraphics[width=\columnwidth]{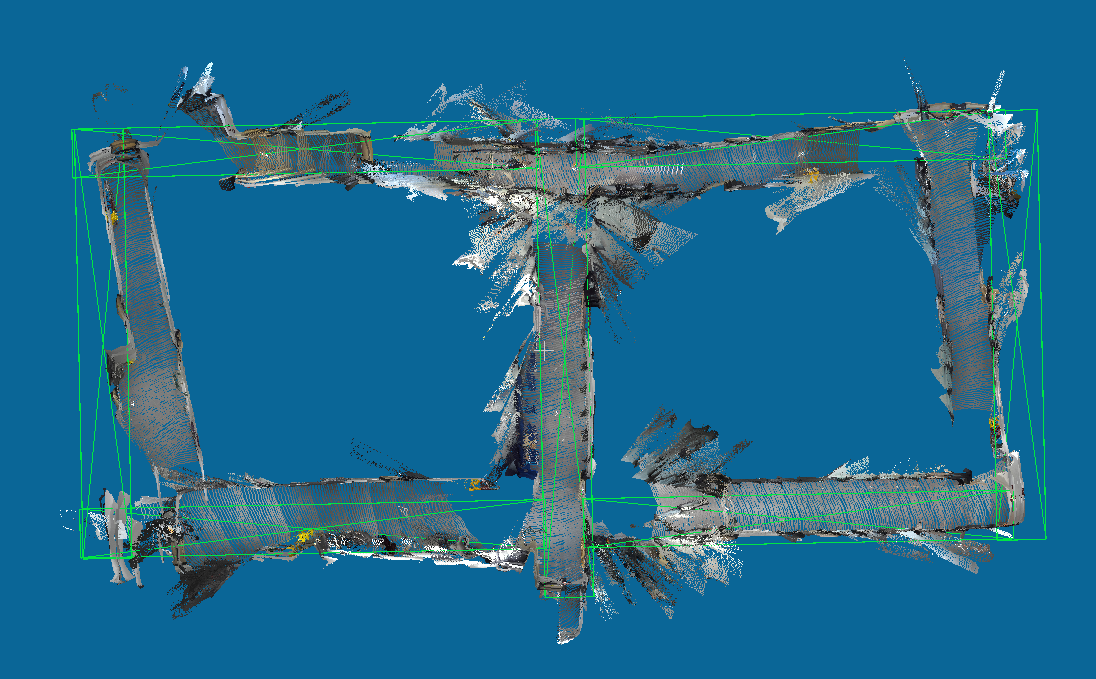}
        \caption{Map with high angular velocity turns overlaid with ground truth wire frame.}
        \label{fig:placeholder}  
    \end{subfigure}

    \caption{Comparison of map generation in an indoor office office environment with differences in turn velocity. Higher turn velocity yielded more drift in the 90 degree angles of the hallway, though the map remains relatively accurate to the ground truth. The slow turns resulted in a high quality reconstruction. These results demonstrate that the external IMU greatly improves the orientation in the state estimate.}
    \label{fig:office_maps}
\end{figure}

\begin{figure}[ht]
    \centering
    \vspace*{+3mm}
    \includegraphics[width=\columnwidth]{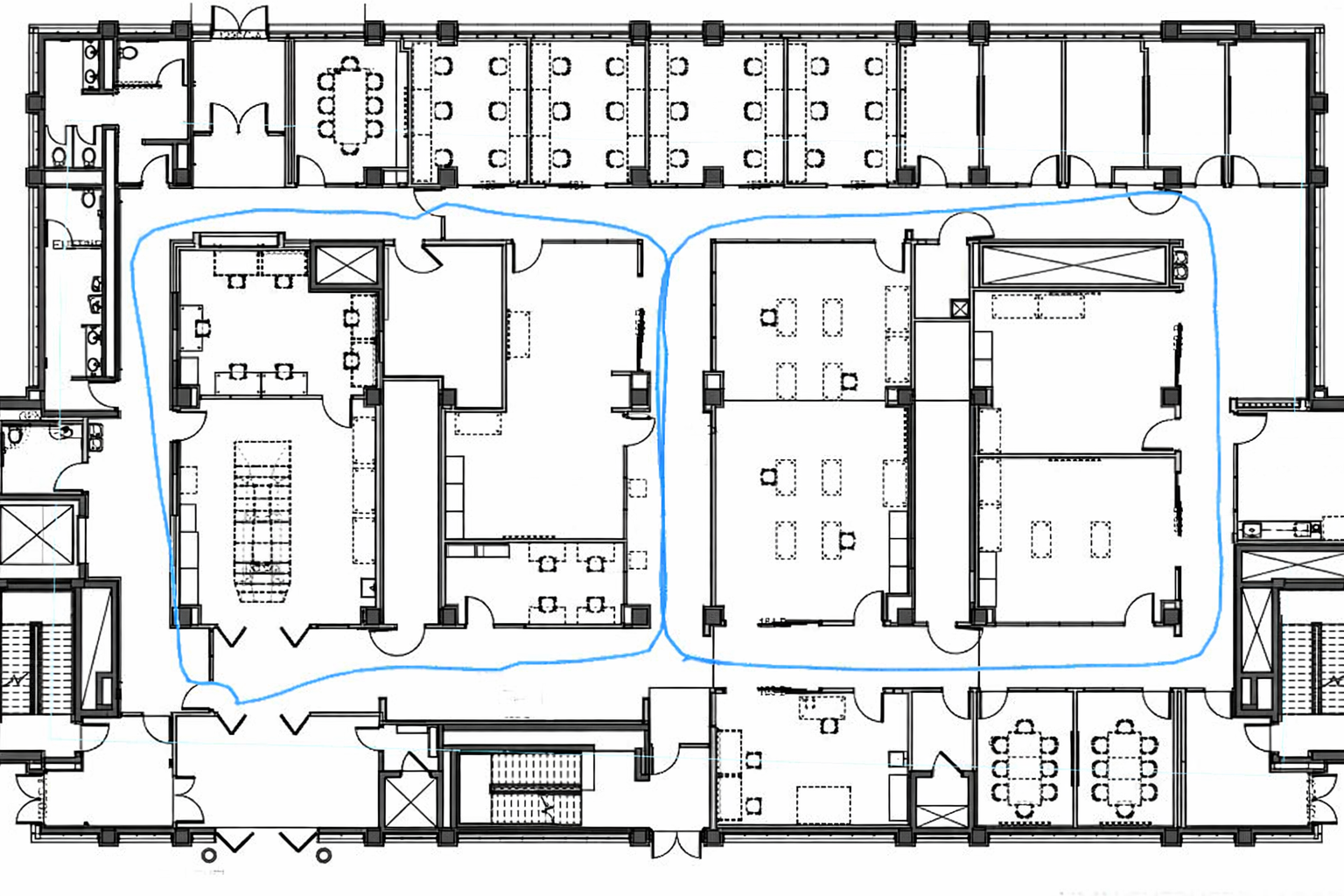}
    \caption{Trajectory of AUV in an indoor office office environment overlaid with a ground truth floor-plan. The trajectory is accurate without any constraints imposed on the assumed movement of the AUV. Notably, the depth sensor keeps the state estimate grounded, preventing vertical drift and ensuring the path is aligned to the 2D floor plan.}
    \label{fig:shep_odom}
\end{figure}

 This experiment was similar to a wheeled vehicle as the AUV was attached to a cart for steady $Z$ displacement. However, when using odometry in a wheeled vehicle, additional constraints can be added to the IMU to reduce drift rate and improve state estimation accuracy~\cite{hendersonRelativePositionUGVs2008}.
 These were not added to this test as the expected operational conditions of the low-cost AUV are of 6-DOF motion. 
 These results validate the utility of the depth sensor, illustrating how an economical component can substantially enhance system accuracy without the need for artificial motion constraints. 
 The resulting maps generated from this experiment can be seen in Fig.~\ref{fig:office_maps}. 
 An overlay of the odometry on the floor plan can also be seen in Fig.~\ref{fig:shep_odom}.

\subsection{Pool Evaluation}
\label{sec:eval_pool}
The pool environment served as a bridge between the structured office and the unstructured ocean, introducing significant perceptual aliasing through uniform tiling. 
To test the limits of the low-cost sensors in feature-sparse conditions, no artificial landmarks were added. 
The AUV followed a trajectory offset from the pool wall, providing a simple rectangular ground truth to evaluate mapping. 
During the trial, the onboard cameras were oriented directly toward the walls or floor to maximize the challenge of limited visual features. 
Although the AUV’s primary translation was forward, its orientation remained unconstrained to simulate `diver-following' behavior, where forward progress is often coupled with arbitrary changes in heading.

\begin{figure}[ht]

    \vspace*{+3mm}
    \begin{subfigure}{\columnwidth}

        \includegraphics[width=\columnwidth]{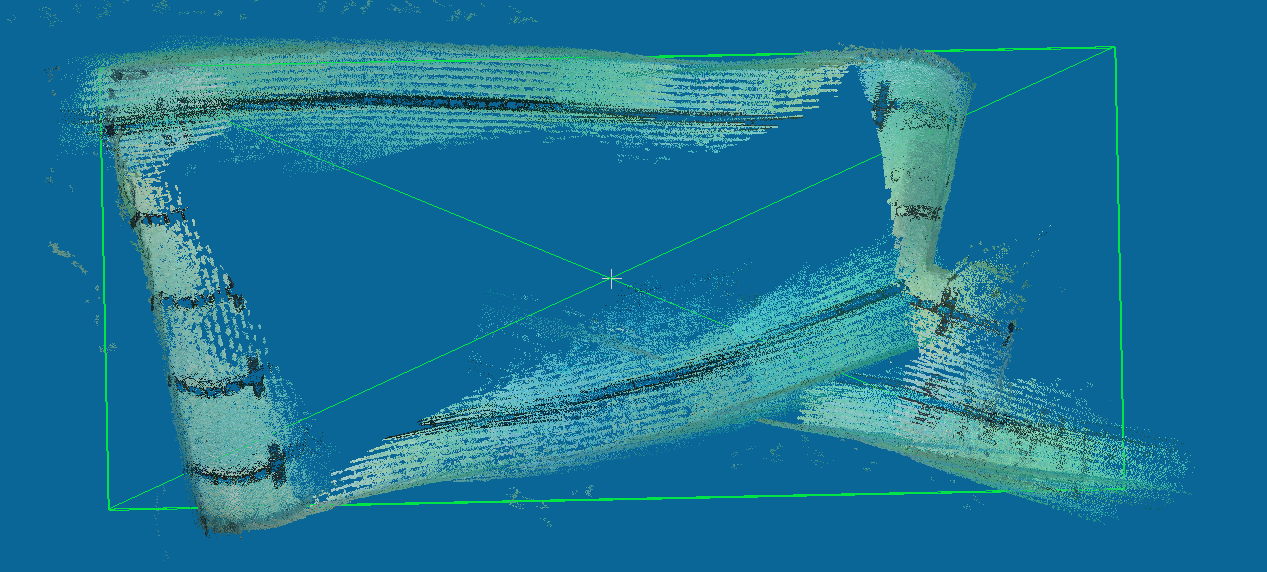}
        \subcaption{Generated pool map with a loop closure overlaid with ground truth wire frame.}
        \label{fig:pool_raw}
    \end{subfigure}
    
    \vspace{.4cm}
    
    \begin{subfigure}{\columnwidth} 

        \includegraphics[width=\columnwidth]{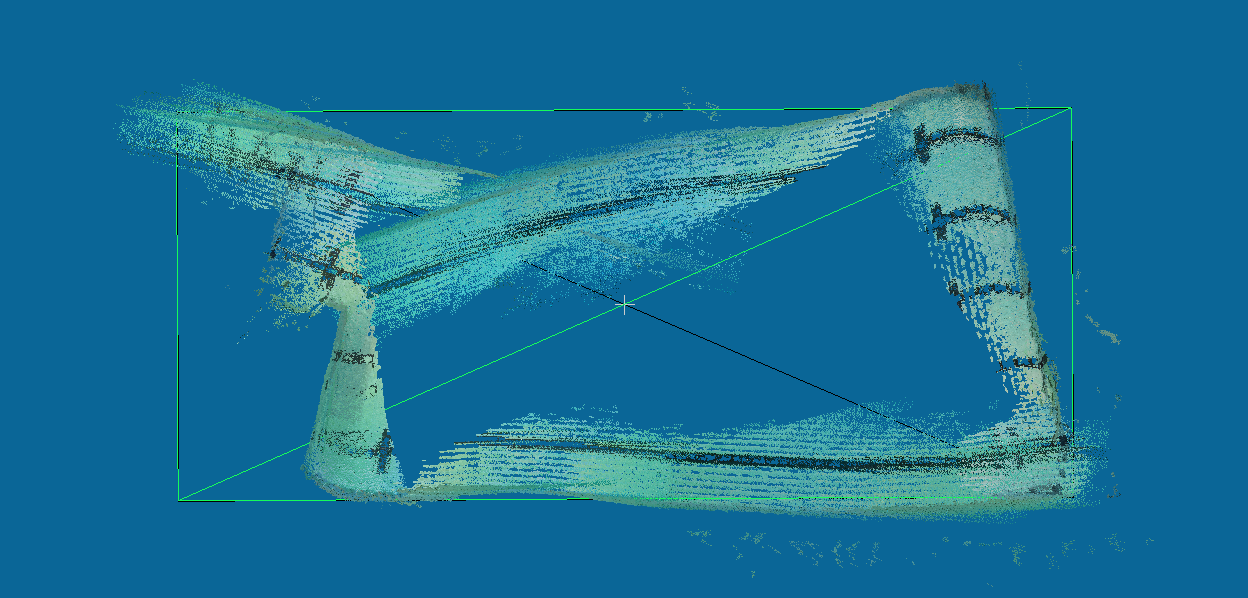}
        \subcaption{Generated pool map without a loop closure overlaid with ground truth wire frame.}
        \label{fig:pool_overlay}
    \end{subfigure}

    \caption{Mapping comparison of pool environment with and without loop closure. With loop closure disabled, a significant gap exists between the start and end position. Loop closure resolves the gap but sacrifices the accuracy of the pool's rectangular geometry. This highlights the inherent challenge of using vision for odometry in environments with repetitive textures.}
    \label{fig:pool_comparison}
\end{figure}

Furthermore, the AUV's vertical displacement was unconstrained throughout this trial. 
This required the system to track depth variations while navigating the pool with its poor visual features. 
This added significant complexity to the state estimation, as the expanded motion profile was coupled with inconsistent light reflections. 
Despite this difficulty, the experiment was designed to have the AUV start and end in the same spot, providing the system a deliberate opportunity for loop closure. 
This scenario pushed the low-cost sensors to their operational limits, providing a benchmark for how much graph optimization can mitigate accumulated drift in adverse underwater conditions. 
See Fig.~\ref{fig:pool_comparison} for results of the mapping.

\subsection{Results and Analysis}
\label{sec:eval_results}

 To assess mapping performance, the physical dimensions of both environments were measured and modeled in a computer Aided Design (CAD) program. 
 These CAD models served as the ground truth meshes for comparison against the generated point clouds using CloudCompare~\cite{CloudCompareOpenSource}. 
 The validation process involved pre-processing the generated maps to remove floor and ceiling points, as the ground truth models were limited to wall geometry. 
 A Statistical Outlier Removal (SOR) filter was then applied to eliminate noise and artifacts. 
 For model synchronization, we employed the Iterative Closest Point (ICP) algorithm to align the point clouds with the ground truth meshes. 
 This alignment facilitated the calculation of the Root Mean Square Error (RMSE) based on point-to-mesh distances. 
 The quantitative results of this analysis are summarized in Table~\ref{tab:rmse_results}.

 \begin{table}[ht]
    \centering
    \caption{RMSE Results of our VIO-SLAM approach from different environments}
    \label{tab:rmse_results}
    \begin{tabular}{l r}
        \toprule
        \textbf{Configuration / Scenario} & \textbf{RMSE Value} \\
        \midrule
        Hallway (Slow Turns)            & $0.56$ m \\
        Hallway (Fast Turns)              & $0.71$ m \\
        Pool With Loop Closure       & $1.04$ m \\
        Pool With No Loop Closure    & $1.14$ m \\
        \bottomrule
    \end{tabular}
\end{table}

The hallway traversal conducted without sharp turns was the most accurate, yielding an RMSE of $0.5$ meters. 
Conversely, the pool mapping without loop closure was the least accurate, with an RMSE of $1.14$ meters.
This performance gap suggests, maybe intuitively, that visual odometry experienced significantly higher drift in the aquatic environment compared to the terrestrial experiment. 
This error likely stems from the photometric consistency assumption used by the ZED’s direct tracking approach. 
This method assumes that the brightness of a point remains constant between frames to calculate motion. 
In the pool, however, dynamic light refractions constantly shift across the floor, violating this assumption and creating ambiguity for the sensors which then perceive incorrect motion.

The performance degradation in the pool was an expected outcome of the experimental design; however, the resulting spatial data remains qualitatively and quantitatively valuable. 
Specifically, the loop-closure mechanism proved robust enough to recover from photometric tracking errors. 
This underscores the effectiveness of our optimization strategy in compensating for the hardware constraints of low-cost sensors. 

\subsection{Ocean Evaluation}
\label{sec:eval_ocean}
\begin{figure}[ht]
    \centering
    \vspace*{+3mm}
    \includegraphics[width=\columnwidth]{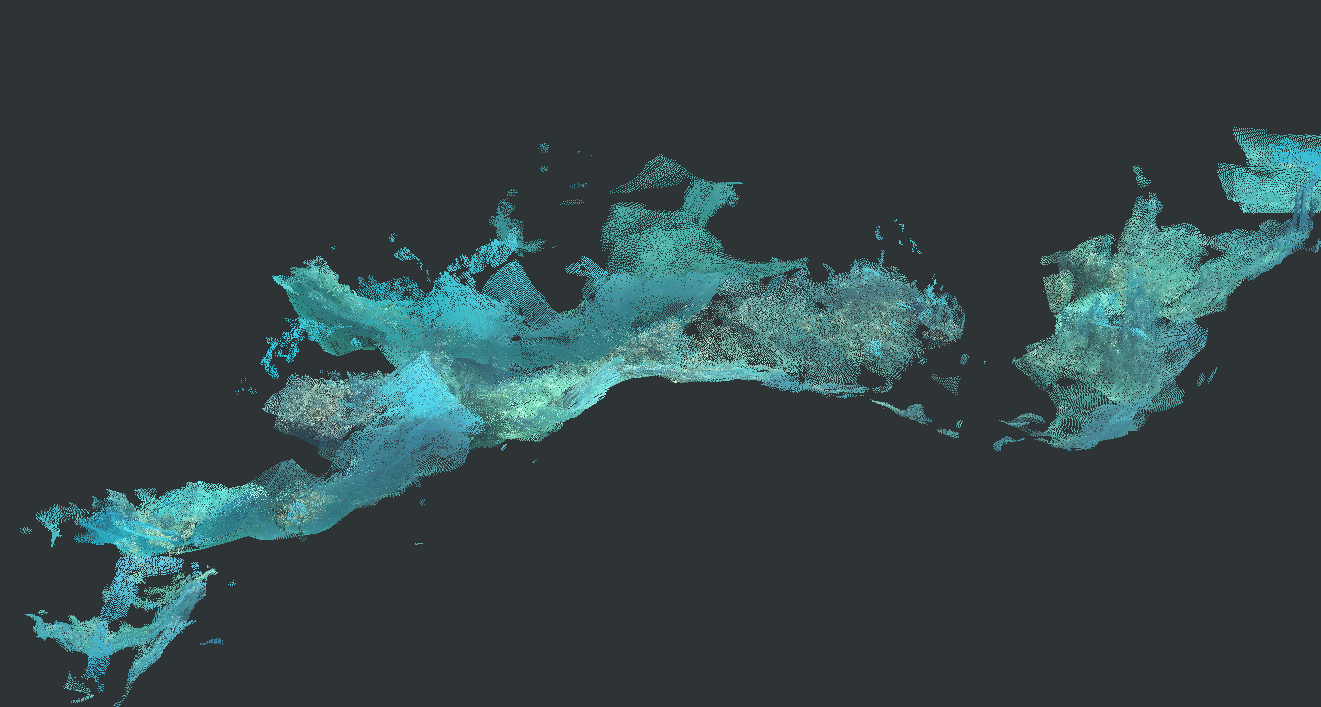}
    \caption{Generated point cloud map of a coral reef. The map shows an accurate path with a distinguishable floor and coral wall. This highlights the ability to obtain quality maps from unconstrained environments, suitable for later analysis of seafloor ecosystems.}
    \label{fig:reef_point_cloud}
\end{figure}

\begin{figure}[ht]
    \centering
    \begin{subfigure}{\columnwidth}
        \centering
        \includegraphics[width=\linewidth]{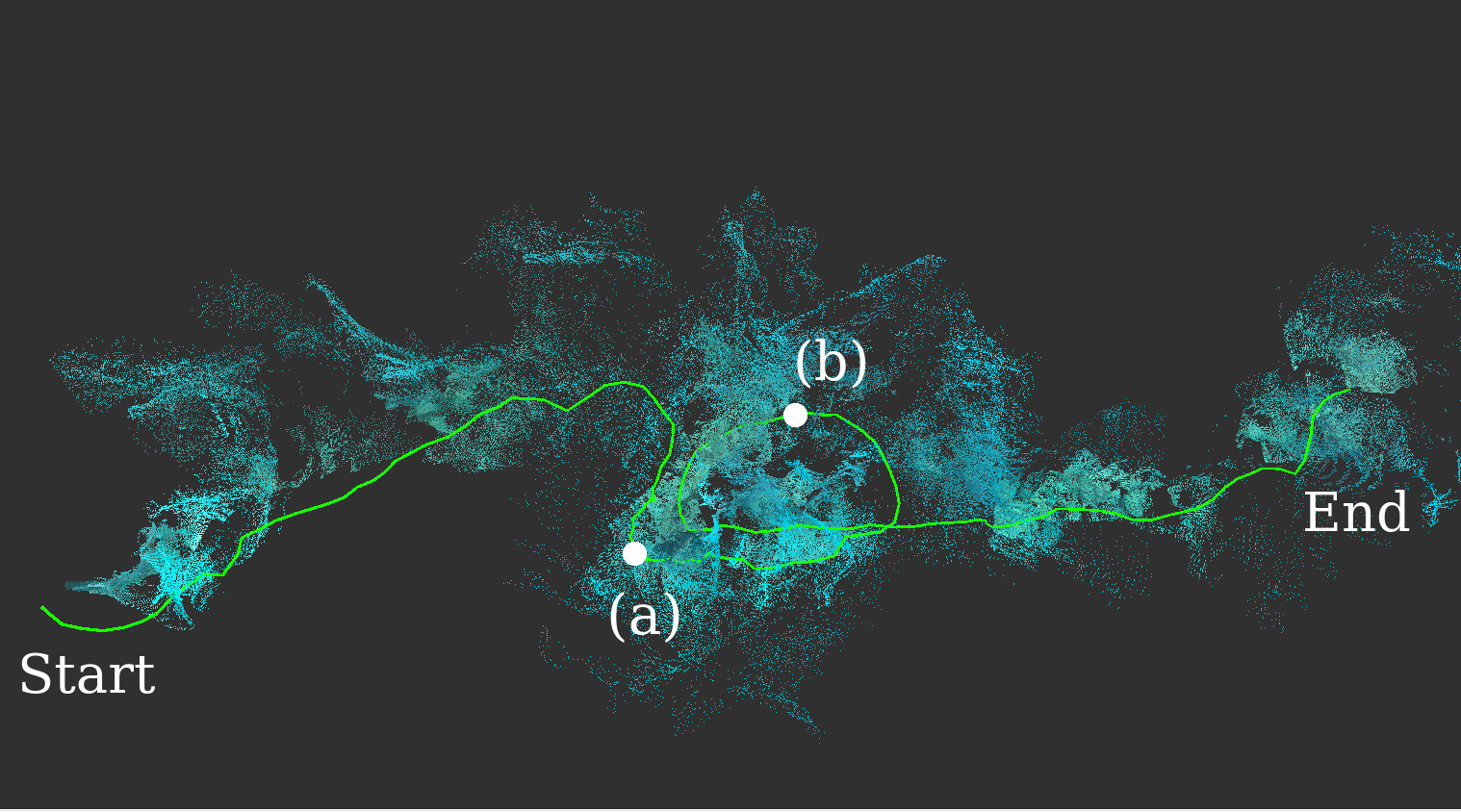}
        \label{fig:main_odom}
    \end{subfigure}

    \begin{subfigure}{0.48\columnwidth}
        \centering
        \includegraphics[width=\linewidth]{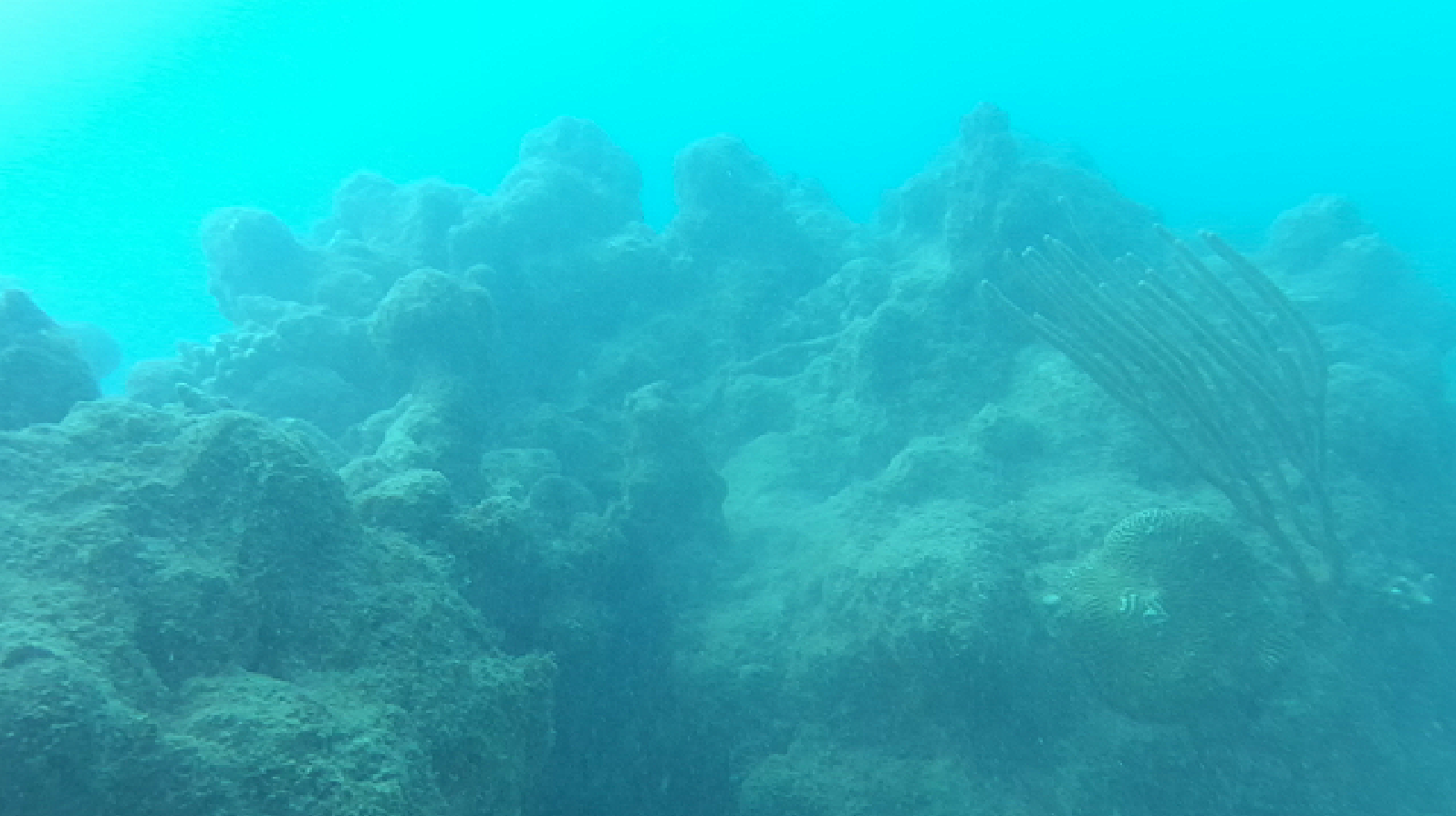}
        \caption{
        \label{fig:sub_left}}
    \end{subfigure}
    \hfill
    \begin{subfigure}{0.48\columnwidth}
        \centering
        \includegraphics[width=\linewidth]{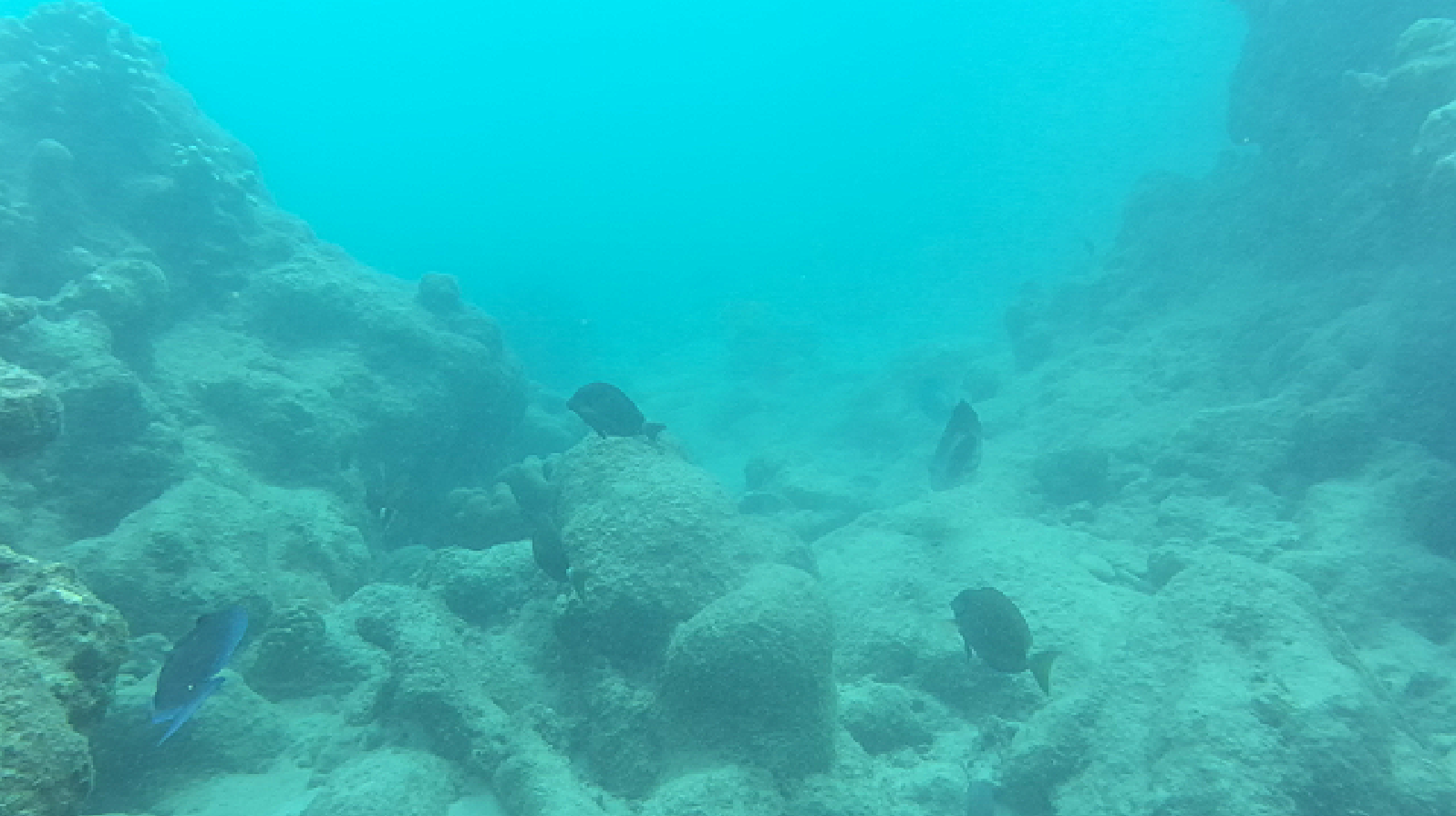}
        \caption{
        \label{fig:sub_right}}
    \end{subfigure}

    \caption{Continuous pose estimation during ocean navigation (top), containing images from the traversed path (bottom). The overlaid path illustrates a smooth state estimate despite environmental constraints. This lack of discontinuities underscores the robustness of the integrated sensor configuration in handling non-linear dynamic data.}
    \label{fig:barbados_odom}
\end{figure}

The final experiment was a deployment in an open-ocean field environment. 
Maps of a coral reef were developed, and while no ground truth exists for this experiment, given a qualitative comparison of the images and generated point cloud (Fig.~\ref{fig:reef_point_cloud} and Fig.~\ref{fig:barbados_odom}), the mapping and odometry were found to be effective enough for vehicle planning and control, even in shallow water with rapidly changing lighting conditions. In Fig.~\ref{fig:barbados_odom}, the images: Fig.~\ref{fig:sub_left} and ~\ref{fig:sub_right} are from the points (a) and (b) labeled on the green path which show some dynamic elements and visibility drop-off. It should be noted however, that in Fig.~\ref{fig:barbados_odom} it can be seen that even with turns, loops, elevation changes and path crossings in a dynamic environment over a coral reef, the AUV is able to create a coherent map of the reef. This demonstrates the utility of this system in real-world environments.

\section{Future Work}
\label{sec:future_work}
We are exploring opportunities to improve the SLAM system aboard the low-cost AUV. 
First, moving away from the StereoLabs VIO may provide benefits. 
The dense implementation, while effective, is computationally expensive. 
Having a depth map calculated for every published camera frame uses a majority of the onboard computing resources (up to $77$\%) of a Jetson Orin NX 16 GPU~\cite{PositionalTrackingModes}. 
Additionally, the ZED SDK is propriety software, so internal changes are not possible. 
We are currently investigating other Open-Source VIO approaches as well as Direct Sparse Approaches~\cite{engelDirectSparseOdometry2018}, similar to Stereo DSO~\cite{wangStereoDSOLargeScale2017} as a viable alternative.

Second, an investigation into sensor fusion methods is underway. 
The EKF while well-studied and effective, is limited. 
Moving Horizon Estimation may lead to better utilization of all the available information. 
This technique allows the incorporation of constraints on the estimated variables. 
It can also perform better on non-linear dynamic systems while being computationally effective, making it a valid avenue to explore~\cite{tennyEfficientMovingHorizon2002}.

\section{Conclusion}
\label{sec:conclusion}
In this work, we have shown that by using only passive sensors, we could successfully deploy a VI-SLAM system on board a low-cost AUV that can run in real time without a tether. 
This system is accurate enough to map a coral reef during operations and provide feedback for an autonomous controller. 
We demonstrated the efficacy of this system in both pool and ocean environments with highly accurate performance. 
We also presented the effect different configurations of the VSLAM system have on the accuracy with a fully unconstrained AUV in open-water conditions.
This work demonstrates that a lower cost AUV can still provide capabilities of far more expensive systems, ensuring accessibility to underwater robotics researchers and users alike.

\bibliographystyle{ieeetr}
\bibliography{Bibliography}
\end{document}